\documentclass[12pt,journal,compsoc, a4paper]{IEEEtran}

\usepackage{amsmath,amsfonts}
\usepackage{mathtools}
\usepackage{algorithmic}
\usepackage[top=2.5cm,bottom=2.5cm,left=1cm,right=1cm]{geometry}
\usepackage{array}
\usepackage{caption}
\usepackage[caption=false,font=normalsize,labelfont=sf,textfont=sf]{subfig}
\usepackage{fancyhdr} 
\usepackage{multirow}
\usepackage{color,soul}
\usepackage[table,xcdraw]{xcolor}
\usepackage{graphicx}
\usepackage{booktabs}
\usepackage{tabularx}
\usepackage{textcomp}
\newcolumntype{Y}{>{\centering\arraybackslash}X}
\usepackage{balance}

\usepackage{booktabs}
\usepackage{listings}
\usepackage{url}
\usepackage{stfloats}
\usepackage[labelfont={bf,footnotesize}, font=footnotesize]{caption}
\usepackage{tikz}
\usepackage{microtype}
\usepackage[hidelinks]{hyperref}
\definecolor{myblue}{rgb}{0, 0.25, 0.45}
\hypersetup{colorlinks=true , 
    linkcolor=.,
    anchorcolor=.,
    citecolor=myblue,
    filecolor=.,
    menucolor=.,
    runcolor=.,
    urlcolor=.}

\usepackage[
    backend=biber,
    style=apa,
    citestyle=authoryear,
    url=false, 
    doi=false,
    natbib=true,
    hyperref=true,
    eprint=false,
    mincitenames=1,
    maxcitenames=2,
    uniquelist=true
]{biblatex}
\renewcommand{\cite}{\parencite}

\def\BibTeX{{\rm B\kern-.05em{\sc i\kern-.025em b}\kern-.08em
    T\kern-.1667em\lower.7ex\hbox{E}\kern-.125emX}}

\DeclareCiteCommand{\citenum}
  {}
  {\printfield{labelnumber}}
  {}
  {}
\definecolor{toqua}{rgb}{0.17,0.3,0.58}
\renewcommand{\headrulewidth}{1pt}
\renewcommand{\headrule}{\hbox to\headwidth{%
    \color{toqua}\leaders\hrule height \headrulewidth\hfill}}

\renewcommand{\footrule}{\hbox to\headwidth{%
    \color{toqua}\leaders\hrule height \headrulewidth\hfill}}
\fancypagestyle{plain}{%
    \fancyfoot[L]{\footnotesize Corresponding authors: \texttt{simon@toqua.ai}, \texttt{casimir@toqua.ai}}
    \fancyfoot[R]{\thepage}
    \fancyhead[C]{\emph{arXiv} Preprint}
    \fancyhead[R]{\today}
  \renewcommand{\headrulewidth}{1pt}
}

\begin{document}
\title{\raggedright Towards Improved Prediction of Ship Performance: A Comparative Analysis on In-service Ship Monitoring Data for Modeling the Speed--Power Relation}

\author{\raggedright
    \IEEEauthorblockN{\href{https://orcid.org/0000-0002-1322-7621}{\includegraphics[scale=0.1]{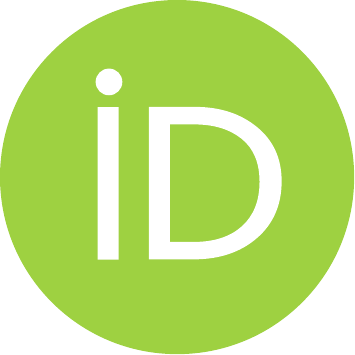}\hspace{1mm}\normalsize Simon DeKeyser}\IEEEauthorrefmark{1}, \normalsize Casimir Morob\'{e}\IEEEauthorrefmark{1}, \href{https://orcid.org/0000-0002-0529-0962}{\includegraphics[scale=0.1]{orcid.pdf}\hspace{1mm}\normalsize Malte Mittendorf}\IEEEauthorrefmark{2}}\\ \vspace{0.5cm}
    \IEEEauthorblockA{\IEEEauthorrefmark{1}\small \emph{Toqua -- Ghent, Belgium}}\\
    \IEEEauthorblockA{\IEEEauthorrefmark{2}\emph{Technical University of Denmark -- Kgs. Lyngby, Denmark} }
}

\IEEEtitleabstractindextext{%
\begin{abstract}
Accurate modeling of ship performance is crucial for the shipping industry to optimize fuel consumption and subsequently reduce emissions. However, predicting the speed-power relation in real-world conditions remains a challenge. In this study, we used in-service monitoring data from multiple vessels with different hull shapes to compare the accuracy of data-driven machine learning (ML) algorithms to traditional methods for assessing ship performance. Our analysis consists of two main parts: (1) a comparison of sea trial curves with calm-water curves fitted on operational data, and (2) a benchmark of multiple added wave resistance theories with an ML-based approach. Our results showed that a simple neural network outperformed established semi-empirical formulas following first principles. The neural network only required operational data as input, while the traditional methods required extensive ship particulars that are often unavailable. These findings suggest that data-driven algorithms may be more effective for predicting ship performance in practical applications.
\end{abstract}
}

\maketitle\thispagestyle{plain}
\begin{tikzpicture}[remember picture,overlay]
\node[anchor=north west,yshift=-15pt,xshift=20pt]%
    at (current page.north west)
    {\includegraphics[height=2 cm]{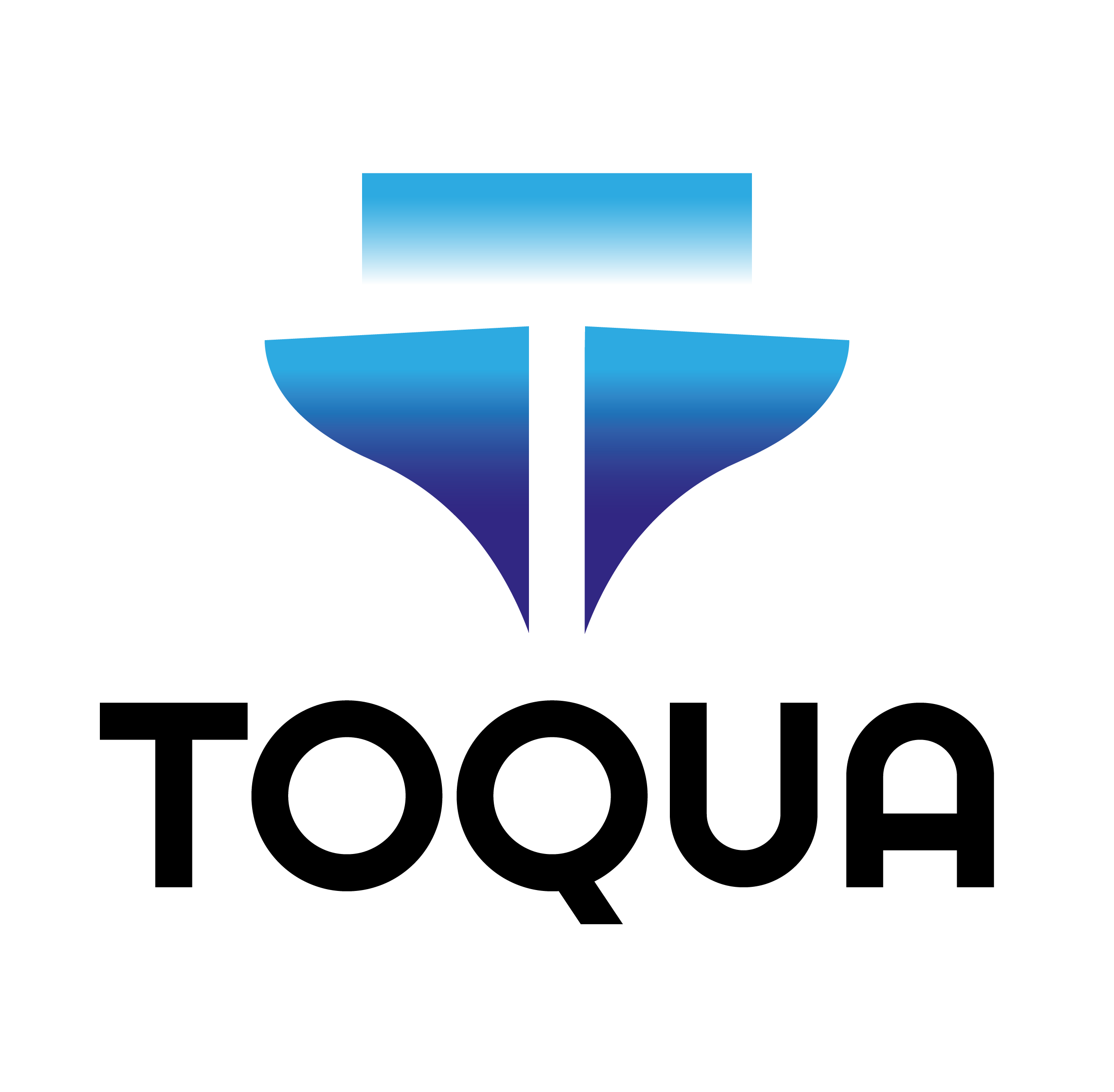}};
\end{tikzpicture}
\vspace{-1cm}
\IEEEpeerreviewmaketitle
\IEEEdisplaynontitleabstractindextext

\section{Introduction}
\IEEEPARstart{A}{ccurate} ship performance modeling is a powerful decision-making instrument for the shipping industry, offering the ability to save fuel and reduce emissions \cite{toqua}, e.g., through optimized hull maintenance and voyage optimization. In the last decades, a spectrum of strategies has been developed, ranging from simplified empirical formulas derived from model tests to advanced full-scale 3-D numerical simulations \cite{tezdogan2015full}. However, considering both accuracy and computational effort, Liu and Papanikolaou \parencite*{liu2020regression} argue that for practical applications where only a limited amount of ship parameters is available, a semi-empirical formula seems to be the most efficient method that captures the underlying physics of the problem. Unfortunately, the development of such methods leads to a loss of accuracy, which we classify into three categories.
\\\\
\textbf{(1)} Approximations are made to generalize the actual hull shape to an efficient number of parameters, i.e., main particulars. The hull form and its displacement distribution significantly impact the magnitude of the wave resistance in calm water. Nevertheless, incorporating the full hull geometry into a semi-empirical model would reduce its practical advantage over full-scale numerical simulations.
\\\\
\textbf{(2)} Assumptions about the range of validity need to be made to enforce regimes where certain physical relations hold. Extrapolating results outside particular validity ranges may result in poor accuracy and non-physical transitions. 
\\\\
\textbf{(3)} The theoretical expressions are fitted on experimental and numerical data obtained from controlled environments, which are far from reality. Most of the experiments are performed on scaled models in towing tanks, causing an increase in errors when scaling the measurements back to full size. The problem lies in a lack of Reynolds number similarity between the ship and the model as only Froude number similarity is obtained. There, further assumptions are made on, e.g., hull smoothness and weather conditions, as imitating every possible ship/environment would be infeasible. Hence, we face data scarcity when assessing the in-service ship performance.
\\\\
The overarching goal of this article is to show that machine learning (ML) approaches may complement or possibly surpass traditional methods in the assessment of in-service ship performance. ML is a subfield of artificial intelligence that involves the use of algorithms and statistical models to enable a system to learn from data and improve its performance on a specific task. It consists of a set of computational techniques to discover patterns and relationships in data and make predictions or decisions based on those patterns. ML algorithms can be thought of as an extension of statistical regression analysis, which is a technique used to model the relationship between a dependent variable (the outcome or response of interest) and one or more independent variables (the predictors or explanatory variables). 
\\\\
Whereas regression analysis typically involves fitting a fixed model to data and then using the model to make predictions or inferences about the data, ML algorithms can be trained on data and adapt as new data becomes available, allowing the system to learn and improve over time without being explicitly programmed for each task. 
\\\\
In the past, naval engineers have pursued finding universal semi-empirical formulas that rely on ship-specific inputs by performing fits (regression analysis) on experimental data to model ship performance. Instead, we propose training ship-specific ML models on real-world data while ensuring physical relevance without needing ship particulars. Nowadays, a lot of in-service sensor data is acquired, allowing us to leverage the scalability of ML and accurately model at once all of the ship-specific intricacies that semi-empirical formulas cannot. 
\\\\
ML is a natural progression for the shipping industry's practical purposes, as it allows us to predict unobserved outcomes in high-dimensional parameter spaces. In fact, several studies have examined the use of ML in the maritime industry, such as fuel consumption prediction and optimization for Diesel engines \cite{parlak2006application}, condition-based maintenance of naval propulsion plants \cite{coraddu2016machine}, and modeling marine fouling speed loss \cite{coraddu2019data, gupta2022ship}. 
\\\\
In this work, the focus lies on predicting the power delivered by the engine as a function of the ship's speed. This so-called speed-power relationship forms the basis of ship performance modeling as it is the gateway to crucial economic parameters such as fuel consumption. An assessment of the current state of ship performance modeling research is made by comparing the accuracy of older, simpler theories with state-of-the-art (semi-)empirical formulas on in-service monitoring data. Herein, engine power sensor measurements will form our ground truth. We follow this approach to enable the extension to a supervised learning problem, where a learning algorithm is trained on observed data (ship speed, weather, and loading conditions) to predict specific targets (engine power) in unobserved situations. 

\section{Methodology}
In the following sections, we use the ship resistance model according to the ISO 19030 \parencite*{ISO19} industry standard to model the speed-power relationship. This model splits the resistance experienced by a ship into different parts:
\begin{equation}
    R = R_{calm} + R_{AA} + R_{AW} + R_{AH} + R_{others}
\end{equation}
where $R_{calm}$ is the calm-water resistance, $R_{AA}$ and $R_{AW}$ respectively are the added resistances due to wind and waves, $R_{AH}$ is the added resistance due to changes in hull condition such as the accumulation of marine growth on hull and propeller, and $R_{others}$ combines the effect of all other contributions such as steering and shallow water resistance.
\\\\
Several (semi-)empirical techniques are elaborated on and critically examined for the different resistance contributions in section \ref{sec:model}. To simplify our analysis, we will discuss the calculation of the first three resistance terms, which constitute the most considerable contribution to $R$ \cite{dalheim2020added}. The other terms require knowledge of ship-specific and environmental parameters, which are often unavailable in real-world situations. 
\\
In this study, we select neural networks (NNs) to prove the value of ML algorithms in predicting ship performance. We chose NNs for their simplicity and effectiveness in demonstrating the potential of ML without adding unnecessary complexity to the comparison. In section \ref{sec:nn}, we provide a brief overview of the NN architecture, pre-processing techniques tailored for in-service vessel data, and the training and evaluation process.
\\\\
Section \ref{sec:results} discusses a benchmark of different $R_{AW}$ estimation procedures evaluated on in-service ship monitoring data to highlight the uncertainties of using experimentally fitted formulas for practical purposes. The benchmark, focusing on the resistance due to calm water, wind, and waves, will provide a clear and functional assessment of the accuracy of the (semi-)empirical techniques discussed in section \ref{sec:model}. 
\\
The goodness of fit between the actual $P_i$ and predicted $\hat{P}_i$ engine powers (for the observations $i = 1, ..., N$) is assessed with several metrics. We use the well-known mean absolute error (MAE):
\begin{equation}
    \text{MAE} = \frac{1}{N} \sum_{i=1}^N \left |P_i - \hat{P}_i\right |
    \label{MAE}
\end{equation}
, the mean absolute percentage error (MAPE):
\begin{equation}
    \text{MAPE} = \frac{1}{N} \sum_{i=1}^N \left | \frac{P_i - \hat{P}_i}{\hat{P}_i} \right |
\end{equation}
, the mean bias error (MBE):
\begin{equation}
    \text{MBE} = \frac{1}{N} \sum_{i=1}^N \left ( \hat{P}_i - P_i \right )
\end{equation}
, and the R-squared score (R2):
\begin{equation}
    \text{R2} = 1 - \sum_{i=1}^N \frac{(P_i - \hat{P}_i)^2}{(P_i - \bar{P})^2}
\end{equation}
with $\bar{P} = \frac{1}{N} \sum_{i=1}^N P_i$ the mean observed power.
\\\\
The in-service dataset combines high-frequency sensor measurements of ship speed and main engine brake power\footnote{The brake power of a ship's engine is the power output of the engine measured at the engine's crankshaft, before any transmission losses. It can be calculated by measuring the torque and angular speed of the crankshaft.} with weather data and loading conditions. Table \ref{tab:variables} summarizes the variables into input and target categories.
\begin{table}[]
\centering
\caption{Input and target variables of the in-service dataset.}
\label{tab:variables}
\begin{tabular}{@{}cl@{}}
\toprule
Category & Variables                                             \\ \midrule
    & Draft aft, Draft forward, Displacement \\
    & Wind direction, Wind speed                            \\
Input    & Wave direction, Significant wave height, Wave period  \\
    & Current speed, Current direction                      \\
    & Speed-through-water, Ship heading       \\ \midrule
Target   & Brake power                                           \\ \bottomrule
\end{tabular}
\end{table}
The data is filtered to \emph{steady state} voyages in full sea conditions, where ship accelerations, shallow water effects, and steering maneuvers are eliminated. 
\\
To further isolate our analysis to the $R_{calm}$, $R_{AA}$ and $R_{AW}$ resistance terms, measurements are taken from periods after cleaning events, such that $R_{AH}$ can be disregarded. Note that the relative contribution of $R_{AH}$ to $R$ will increase over time and for practical use cases, accurately modeling $R_{AH}$ is crucial. Although it remains a challenging task to solve using empirical models \cite{guo2022combined, demirel2017effect}, recent research has shown that digital twin ML-based methods may bring solace \cite{coraddu2019data, gupta2022ship}.
\\\\
Finally, a comparison is made with a simple ship-specific NN model, which incorporates all of the resistance contributions while requiring only operational data as input. Here, we also consider some limitations of using ML for ship performance modeling. Section \ref{sec:conclusion} will summarize our findings and use them to propose ML as the natural next step towards improved prediction of ship performance.

\section{Ship Resistance Model}
\label{sec:model}
The brake power $P_B$ of the ship's main engine is calculated according to:
\begin{equation}
    P_B = \frac{R V_S}{\eta_D \eta_M}
    \label{eq:Pe}
\end{equation}
with $V_S$ the ship's speed-through-water (STW), $\eta_D$ the propulsive efficiency, $\eta_M$ the mechanical (shaft  + gearbox) efficiency, and $R$ the total resistance. As with $R_{AH}$, empirically modeling the propulsive and mechanical efficiencies with limited information remains difficult \cite{shigunov2017added}. 
\\
In particular, $\eta_D$ depends on the wake fraction, thrust deduction, propeller diameter, and even the total resistance experienced by the hull. All these factors but the propeller diameter are speed and seaway dependent. Several curves and empirical models have been proposed to estimate $\eta_D$ \cite{fluid_mechanics_1995, kristensen2012prediction, holtrop1982approximate}. However, these are only valid for calm water, and there is a lack of data in waves. Therefore, we use the default values recommended by ISO19030.

\subsection{Calm-Water Resistance}
A direct method for calculating a ship's calm-water resistance is to conduct a sea trial, during which the speed-power relation is measured under calm weather conditions. Usually, these measurements are performed around the design speed of the vessel. In recent years, however, operational velocities have been lowered to reduce emissions \cite{psaraftis2014ship}, meaning that operating speeds do not always match the design speeds anymore. 
\\
It is widely accepted that the correlation between speed and power follows a relation $P \approx V^c$ with constant exponent $c \approx 3$, which holds around the design speed \cite{solutions2018basic}. However, previous studies have found that the \emph{cubic law} underestimates power at speeds below the design speed \cite{adland2020optimal, taskar2020benefit, tillig2018analysis}. These findings suggest that simply extrapolating sea trial curves to operational speeds lower than the design speed can lead to underestimations of power.
\\\\ 
In this work, we will omit three of the main issues associated with the use of sea trial curves to estimate $R_{calm}$:
\begin{itemize}
    \item Sea trial curve availability
    \item Operational speed $\neq$ design speed
    \item Time-dependent performance loss (e.g., fouling)
\end{itemize}
by using the data-driven method proposed by Berthelsen and Nielsen \parencite*{berthelsen2021prediction}. This method fits a calm-water curve using in-service data. Their model starts from a simple least-squares fit of ($x_1, x_2$) to a log-linearized power law:
\begin{align}
    P &= x_1V^{x_2}\\
    \ln(P) &= \ln(x_1) + x_2 \ln(V)
\end{align}
They extend the regression model to incorporate the ship's draft $T$:
\begin{equation}
    \ln(P) = \ln(x_1) + x_2 \ln(V) + x_3 T+  x_4 \ln(V)T
\end{equation}
and make the exponent speed dependent by introducing breakpoints that separate different speed intervals where the exponent remains constant. With one breakpoint $B_p$, the regression is performed on:
\begin{align}
\begin{split}
    \ln(P) = \ln(x_1) &+ x_2 \ln(V) + x_3 T+  x_4 \ln(V)T \\
    &+ x_5 (\ln(V) - \ln(B_p))V_d
\end{split}
    \label{eq:calm}
\end{align}
In their paper, $V_d$ is a Heaviside function centered at $B_p$. We, however, propose a differentiable dummy index to make the speed-power relation smooth:
\begin{equation}
    V_d = \frac{1}{2}\left(1 + \tanh\left(\frac{V-B_p}{\delta}\right)\right)
\end{equation}
with $\delta$ the smoothing factor. 
\\\\
In practice, the breakpoints are detected with a binary segmentation algorithm, as implemented in the \texttt{ruptures} Python library \cite{truong2020selective}. The algorithm is applied to the measured power of the speed-sorted data, and it finds change points in the signal where the slope of speed and power alters. At last, the regression is performed on the power data, which is first corrected from weather conditions (see next sections) to ensure that we fit calm-water data. As we do not correct the fouling power loss, a part of $R_{fouling}$ will be included in the fitted calm-water curves.

\subsection{Added Resistance due to Wind}
\begin{figure}[!t]
\centering
\includegraphics[width=\linewidth]{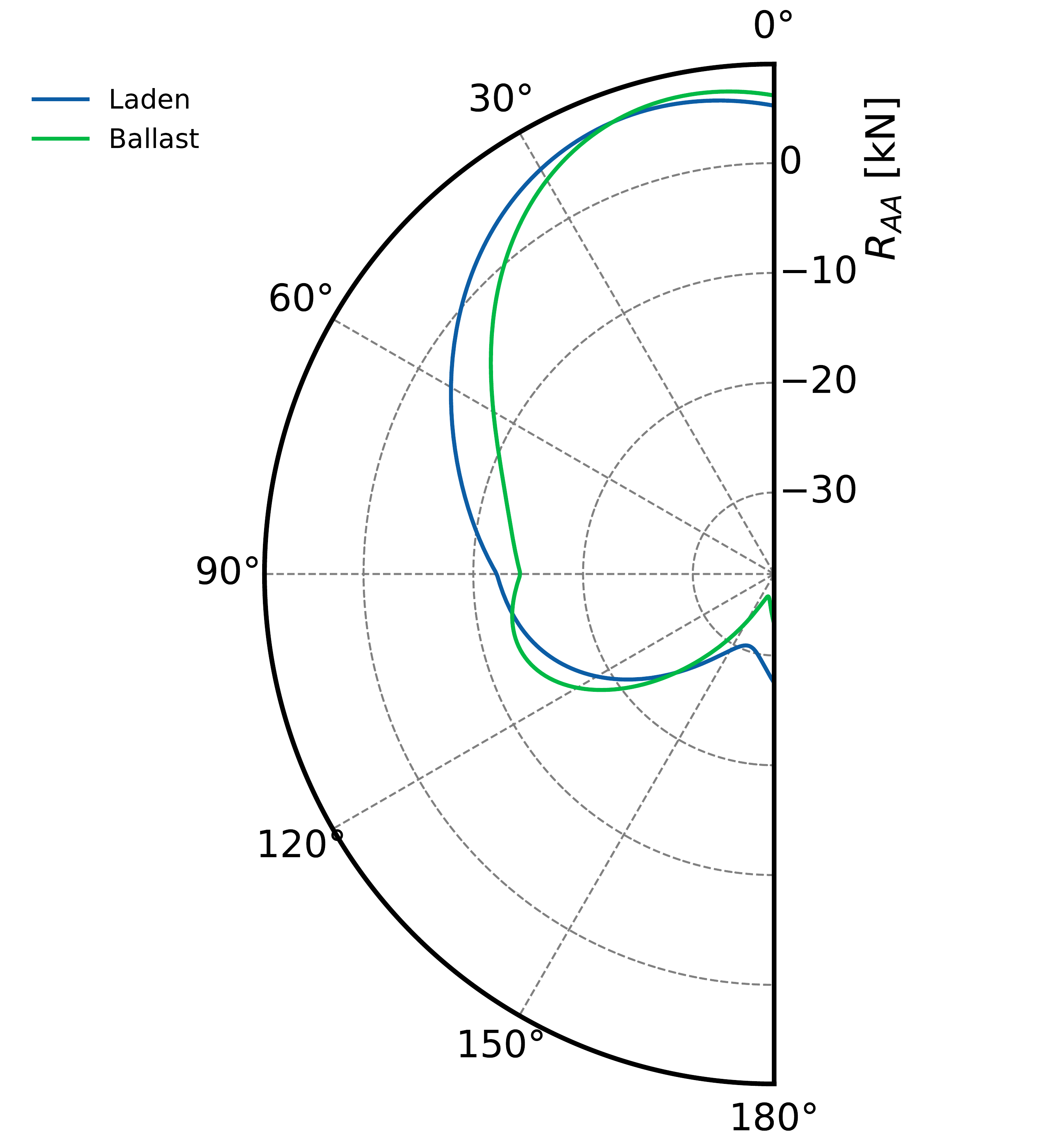}
\caption{Polar plot of the added resistance due to wind as a function of the relative wind direction, calculated with Fujiwara's and Kitamura's regression formulas for different ship load conditions.}
\label{winddrags}
\end{figure}

The industry standard for calculating $R_{AA}$ is: \cite{ISO15}
\begin{equation}
    R_{AA} = \frac{1}{2}\rho_A A_{XV}C_{AA}(\theta_{rel})V_{wrel}^2 - \frac{1}{2}\rho_A A_{XV}C_{AA}(0^{\circ})V_{G}^2
    \label{eq:RAA}
\end{equation}
with $\rho_A$ the air density, $V_{wrel}$ the relative wind speed calculated according to ISO 15016, $V_G$ the ship's measured speed over ground, $A_{XV}$ the transverse projected area of the ship above the water line, $C_{AA}$ the wind resistance coefficient and $\theta_{rel}$ the relative wind direction. The negative term is the air resistance due to the ship moving forward with a headwind caused by $V_G$ ($\theta_{rel} = 0^{\circ}$), which is already included in $R_{calm}$.
\\\\
$C_{AA}$ is estimated with the regression formula based on wind tunnel tests developed by Fujiwara \parencite*{fujiwara2006new}:
\begin{align}
\begin{split}
    C_{AA}(\theta_{rel}) = \: &C_{LF}\cos(\theta_{rel}) + \\
    &C_{XLI}\left(\sin(\theta_{rel}) - \frac{1}{2}\sin(\theta_{rel})\cos^3(\theta_{rel})\right) +\\
    &C_{ALF}\sin(\theta_{rel})\cos^3(\theta_{rel})
\end{split}
\end{align}
where the coefficients $C_{LF}$, $C_{XLI}$, and $C_{ALF}$ consist of different regression expressions for $\theta_{rel} < 90^{\circ}$ and $\theta_{rel} > 90^{\circ}$. Unfortunately, the latter coefficients depend on several detailed ship geometry-related parameters, such as the bridge height and longitudinal projected area of superstructures, which are usually unknown. 
\\
However, Kitamura et al. \parencite*{kitamura2017estimation} developed ship-type-specific regression formulas that estimate the input parameters $P$ of Fujiwara's formula and $A_{XV}$ from the ship's overall length $L_{OA}$ and beam $B$:
\begin{equation}
    \begin{rcases}
        P\\
        P/L_{OA}\\
        P/B\\
        P/L_{OA}^2\\
        P/(L_{OA}B)\\
        P/B^2
    \end{rcases}
    = 
    \begin{cases}
        aB + bL_{OA} + c\\
        aB + c\\
        bL_{OA} + c
    \end{cases}
\end{equation}
with ($a$, $b$, $c$) the regression coefficients and where the left-hand-side and right-hand-side expressions were carefully chosen for each specific parameter to maximize the accuracy. In the following, this approach is used to calculate the inputs for Fujiwara's expression, and while it introduces additional approximations, it is necessary to keep the required parameters at a feasible level.
\\\\
Fig. \ref{winddrags} shows $R_{AA}$ calculated according to this method for a bulk carrier ($L_{OA} = 190$ m, $B = 32$ m) with $V_{wrel} = 8$ m/s and $V_G = V_S = 13$ kn, where a distinction is made between laden and ballast loading conditions. The absolute value of $R_{AA}$ is higher for ballast conditions, which is expected as the area above water is larger. Also, the added resistance is more considerable for headwinds and drops to negative values for following winds, meaning that the ship is being pushed forward. The reader may find it counter-intuitive that $R_{AA} = 0$ kN occurs at angles of only $\pm 40^{\circ}$, but it is stressed that we plot the added resistance relative to the ship's headwind (Eq. \ref{eq:RAA}).

\subsection{Added Resistance due to Waves}
\begin{figure}[!t]
\centering
\includegraphics[width=\linewidth]{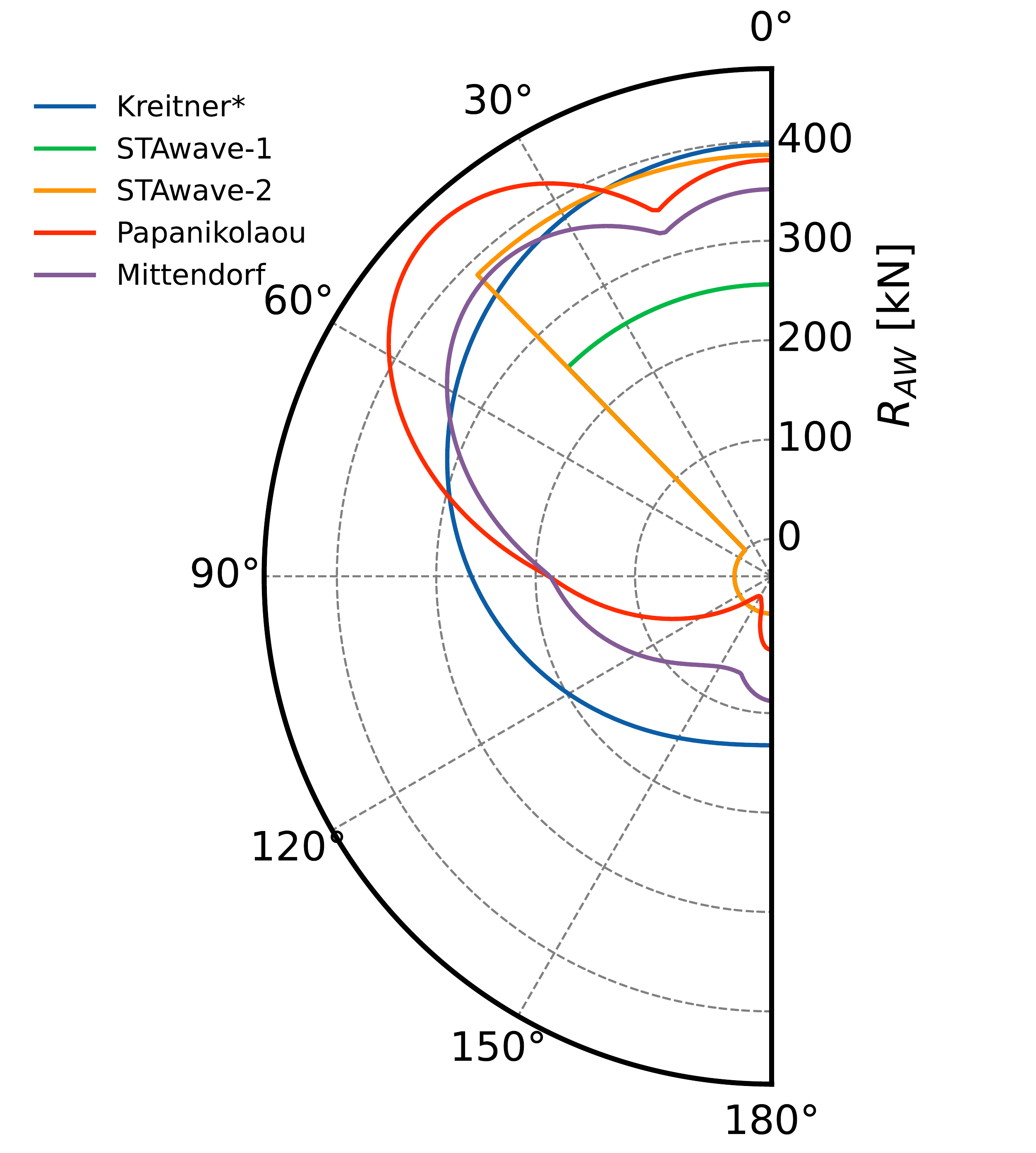}
\caption{Polar plot of the added resistance due to waves as a function of the relative wave direction, calculated with different wave resistance theories in laden condition.}
\label{wavedrags}
\end{figure}

\begin{figure}[!t]
\centering
\includegraphics[width=\linewidth]{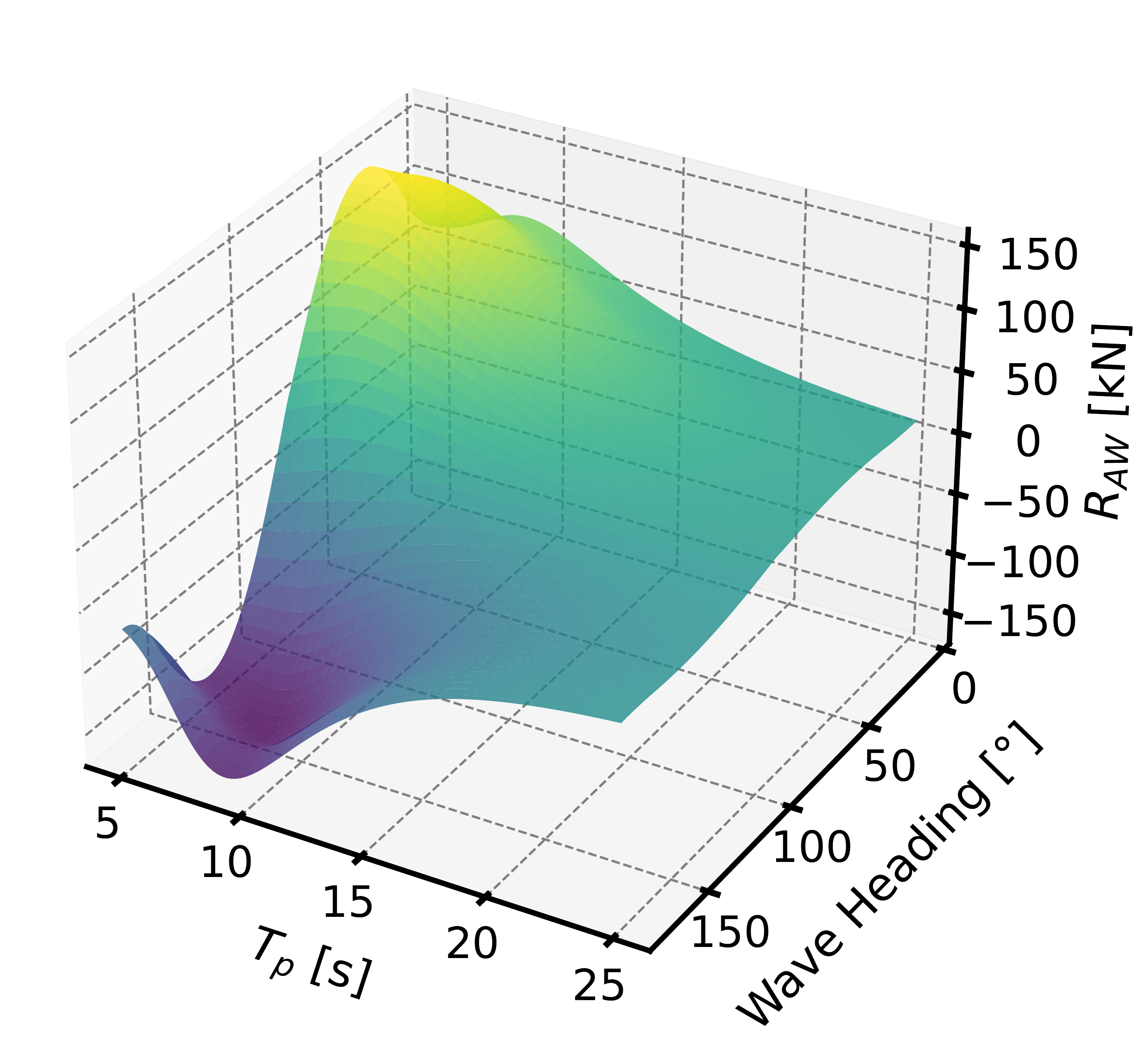}
\caption{Surface plot of $R_{AW}$ as a function of $T_p$ and the wave heading $\alpha_{rel}$ calculated with the Mittendorf method in laden condition.}
\label{peakperiod}
\end{figure}

Modeling a ship's added resistance due to waves is a highly complex, non-linear problem, and most methods rely on simplified assumptions. In practice, $R_{AW}$ is defined as ``the unsteady longitudinal force a ship experiences apart from the calm water, and wind resistances in a realistic seaway'' \cite{mittendorf2022data}. 
\\
The force is a second-order quantity, which depends on the incident wave's amplitude and speed \cite{liu2011prediction}. One of the oldest and simplest modeling approaches is that of Kreitner, used in ITTC2005 \parencite*{ITTC2005}:
\begin{equation}
    R_{AW} = 0.64 g H_S^2 C_B \rho_w \frac{B^2}{L_{pp}}
\end{equation}
with $g$ the gravitational constant, $H_S$ the significant wave height, $C_B$ the block coefficient, $\rho_w$ the water density, and $L_{pp}$ the length between perpendiculars. As this expression only holds for head waves, a cosine law can be used to account for waves with an arbitrary heading \cite{hansen2011performance}:
\begin{equation}
    R_{AW} = 0.64 g H_S^2 C_B \rho_w \frac{B^2}{3L_{OA}}(2 + \cos(\alpha_{rel}))
\end{equation}
with $\alpha_{rel}$ the relative wave direction.\footnote{We use $\alpha_{rel} = 0^{\circ}$ for head waves, and $\alpha_{rel} = 180^{\circ}$ for following waves.} 
\\
Another simple, but more recent standard of ITTC 2014 \cite{ITTC2014}, is the empirical STAwave-1 formula:
\begin{equation}
    R_{AW} = \frac{1}{16} g H_S^2 \rho_w B \sqrt{\frac{B}{L_{B}}} \qquad |\alpha_{rel}| \leq 45^{\circ}
\end{equation}
where $L_{B}$ is the length of the bow at the waterline.
\\\\
More advanced (semi-)empirical approaches split the added wave resistance into two contributions:
\begin{equation}
    R_{AW} = R_{AWM} + R_{AWR}
\end{equation}
where the motion induced $R_{AWM}$ and wave reflection $R_{AWR}$ resistances are a function of the wave frequency $\omega$. To accurately model the ship's resistance in irregular sea conditions with waves of varying frequencies, integration is performed over a wave spectrum $S(\omega)$ to calculate the mean added resistance:
\begin{equation}
    \overline{R}_{AW} = 2\int_0^{\infty} S(\omega) \frac{R_{AW}(\omega)}{\zeta_a^2} d\omega
    \label{eq:int}
\end{equation}
with $\zeta_a$ the wave amplitude. Hereby, we assume that the superposition principle holds and that the calculation is valid for long-crested waves. Often, a Pierson-Moskowitz type spectrum is used in fully developed sea states:
\begin{equation}
    S(\omega) = \frac{5}{16}H_S^2 \frac{\omega_p^4}{\omega^5} \exp\left(-\frac{5}{4}\left(\frac{\omega_p}{\omega}\right)^4\right)
\end{equation}
where $\omega_p = 2\pi/T_p$ and $T_p$ is the wave peak period. 
\\
Using this frequency response framework, ITTC2014 \parencite*{ITTC2014} and ISO 15016 \parencite*{ISO15} recommend using the semi-empirical method called STAwave-2, which again holds for $|\alpha_{rel}| \leq 45^{\circ}$. 
\\\\
From 2016 to 2020, Liu and Papanikolaou \parencite*{liu2016fast, liu2020regression} derived an improved formula with regression analysis on model test data, which holds for arbitrary wave headings. In 2022, Mittendorf et al. \parencite*{mittendorf2022towards} enhanced their formula by performing multivariate regression on the parameter vector with model test data, thereby also including the 90 \% prediction interval of $R_{AW}$. We note that a discontinuity is present in the Liu/Mittendorf formulas, which we found to be caused by the non-linear form factor in the expression for $R_{AWR}$:
\begin{equation}
    \left(\frac{0.87}{C_B}\right)^{(1+4\sqrt{Fr})f(\alpha_{rel})}
\end{equation}
with $Fr$ the Froude number and:
\begin{equation}
f(\alpha_{rel}) = 
    \begin{cases}
        -\cos(\alpha_{rel}) & \pi - E_1 \leq \alpha_{rel} \leq \pi \\
        0  & \alpha_{rel} < \pi - E_1
    \end{cases}
\end{equation}
where $E_1$ is an angle that defines where the ship's bow ends. The discontinuity can be solved by shifting the cosine of $f(\alpha_{rel})$ to 0 when $\alpha_{rel} = \pi - E_1$, while still forcing $f(\alpha_{rel}) = 1$ for $\alpha_{rel} = \pi$:
\begin{equation}
f(\alpha_{rel}) = 
    \begin{cases}
        -\frac{\cos(\alpha_{rel}) + 1}{\cos(\pi - E_1) + 1} + 1 & \pi - E_1 \leq \alpha_{rel} \leq \pi \\
        0  & \alpha_{rel} < \pi - E_1
    \end{cases}
\end{equation}
We will use this new form of $f(\alpha)$ in the $R_{AWR}$ expression of the Liu and Mittendorf methods, as discontinuities will not be present in the measured data.
\\\\
To give the reader a visual understanding of the different wave resistance formulas, Fig. \ref{wavedrags} plots $R_{AW}$ calculated with the aforementioned methods for a bulk carrier ($L_{OA} = 190$ m, $B = 32$ m, $C_B = 0.7$) with $H_S = 4$ m, $T_p = 10$ s and $V_G = V_S = 13$ kn in laden condition. Given Fig. \ref{wavedrags}, it is stated that while the theories' $R_{AW}$ predictions are in the same order of magnitude, their shapes vary differently as a function of $\alpha_{rel}$. 
\mbox{}\\\\
Although the discontinuity has been resolved, the semi-empirical Liu and Mittendorf methods still exhibit non-differentiable irregularities where the resistance increases or decreases. The reason is that their formula consists of expressions for the bow and stern of the ship, including form factors to incorporate non-linear behavior. Of course, in natural seas, the waves will come from multiple directions at once, and Eq. \ref{eq:int} can be extended to a double integration over an angular wave spectrum. Accounting for short-crested waves will smooth the $R_{AW}(\alpha_{rel})$ curves to behave as one would expect physically. Unfortunately, performing a two-dimensional numerical integration comes at the cost of computational time.
\\
Another observation is that the more advanced methods predict a larger wave resistance for oblique waves compared to head waves. The latter effect is confirmed in model tests, where oblique waves cause more pitch motions dissipating energy \cite{valanto2015experimental}. The magnitude of the wave angle-dependent Kreitner's formula remains similar for the different wave headings and lies somewhere between the STAwave and the Mittendorf methods. 
\\\\
Lastly, to highlight the importance of the peak wave period on the frequency response-type methods, Fig. \ref{peakperiod} shows a surface plot of $R_{AW}$ as a function of $T_p$ and the wave heading $\alpha_{rel}$ calculated with the Mittendorf method. The same values were used as in Fig. \ref{wavedrags} for $B$, $H_S$, etc. and $V_G$ = $V_S$ = 0 kn. The ship experiences the highest added resistance in sea states with short waves that arrive obliquely to the ship's heading.

\section{An ML-based Approach}
\label{sec:nn}
Neural networks (NNs) are ML models inspired by the structure and function of the human brain. They are composed of interconnected processing nodes, called neurons, which work together to recognize patterns in data and make predictions. Here, we will use one of the most straightforward NN architectures, the feedforward NN (FNN). This section will briefly explain the architecture of the FNN, the data pre-processing steps, and how a standard training procedure is performed and validated.

\subsection{The Feedforward Architecture}
An FNN is organized into three main parts: the input layer, the hidden layers, and the output layer. The input layer receives input data, the hidden layers process that data and the output layer yields the final output of the network. Each layer consists of several neurons, which receive input from the previous layer and produce the output that is conveyed to the next layer.
\\\\
Mathematically, a neuron can be represented as a function that takes in a set of inputs, $x_1, x_2, ..., x_n$, and produces a single output, $y$. The inputs are multiplied by a set of weights, $w_1, w_2, ..., w_n$, and summed with a bias term, $b$, to produce an intermediate value, $z$, which is then passed through a non-linear activation function, $f$, to produce the output:
\begin{align}
    z &= \sum_{i=1}^{n} w_i x_i + b\\
    y &= f(z)
\end{align}
A commonly used activation function is the rectified linear unit (ReLU):
\begin{equation}
    f(z) = \text{max}(0, z)
\end{equation}
which has the disadvantage of being non-differentiable at $z = 0$. In this work, we use a smooth approximation of the ReLU function, namely the softplus activation function, defined as:
\begin{equation}
    f(z) = \log(1 + e^z)
\end{equation}
The main premise of activation functions is to replicate the behavior of brain neurons, which fire when the electrical impulses coming from connections with other neurons reach a certain threshold (at $z = 0$ here).
\subsection{Feature Engineering}
The input features are the variables presented to the network's input layer. The process of selecting and transforming the input features to improve the performance of a NN is called feature engineering. It involves identifying the most relevant features in the dataset and pre-processing them in a way that makes them more suitable for the NN to learn from. 
\\\\
In this work, we deliberately reduce the NN's complexity by using as few features as possible without losing predictive power by combining different input variables from table \ref{tab:variables}. 
\\
Table \ref{tab:features} summarizes the engineered features, which mimic theoretical formulas of wave and wind resistance. E.g., the \emph{Wind Product} feature resembles the form of Eq. \ref{eq:RAA}, while the \emph{Wave Power} mimics the power per unit wave crest in deep water:
\begin{equation}
    P = \frac{\rho_w g}{64 \pi} H_S^2 T_e
\end{equation}
with $T_e$ the wave energy period. 
\begin{table}[!b]
\centering
\caption{Features used as input of the NN.}
\label{tab:features}
\bgroup
\def\arraystretch{1.2}
{\setlength{\tabcolsep}{15pt}
\begin{tabular}{@{}ll@{}}
\toprule
Feature                  & Formula                         \\ \midrule
STW                      &                                \\
Draft Average            & $(T_{aft}$ + $T_{fwd})/2$             \\
Wind Product (long.)     & $V_{wrel}^2 \cos(\theta_{rel})$ \\
Wind Product (trans.)    & $V_{wrel}^2 \sin(\theta_{rel})$ \\
Wave Power (long.)       & $H_S^2 T_p \cos(\alpha_{rel})$  \\
Wave Power (trans.)      & $H_S^2 T_p \sin(\alpha_{rel})$  \\ \bottomrule
\end{tabular}}
\egroup
\end{table}
\mbox{}\\
Although power is not a vector quantity, we separate all directional
variables into their longitudinal (long.) and transversal (trans.) direction to the vessel's heading. Doing so results in a correct angular dependence of model w.r.t. the wind direction $\theta_{rel}$ and wave direction $\alpha_{rel}$ relative to the vessel's heading. 
\\\\
The final FNN architecture consists of an input layer with six features, four fully-connected hidden layers (64, 32, 16, and 8 neurons, respectively), and an output layer that predicts the brake power. 

\subsection{Training and Validation}
During training, the weights and biases of the neurons are adjusted to minimize a loss function that measures the difference between the predicted output and the true output (e.g., with the MAE (Eq. \ref{MAE}) or mean squared error (MSE)). The optimization is generally conducted with algorithms like stochastic gradient descent, which calculates the gradient of the loss function with respect to the weights and biases and updates the weights and biases in the direction that reduces the loss. 
\\
In this study, we employed ADAM \cite{kingma2014adam}, a frequently used variation of gradient descent that incorporates momentum \cite{rumelhart1985learning} to accelerate convergence to the minimum of the loss function in the high-dimensional space of weights and biases.
\\\\
To assess the performance of the NN, we use k-fold cross-validation. Here, the data is divided into k \emph{folds}, and the model is trained on k-1 of those folds while leaving the remaining fold as a validation set. This process is repeated k times, with a different fold being used as the validation set each time. The final performance of the model is then calculated as the average performance across all k iterations. This approach ensures that the NN is tested on completely unseen data, giving us a reliable representation of how the model would perform in unobserved real-world conditions.


\section{Discussion}
\label{sec:results}
In the following, we will use in-service data of different vessels to compare the performance of the resistance models described above and extend the analysis to include a simple NN. Before proceeding, we will first justify the use of a speed-dependent exponent in the calm-water speed-power relation.
\subsection{Calm-Water Resistance}
\begin{figure}[!t]
\centering
\includegraphics[width=\linewidth]{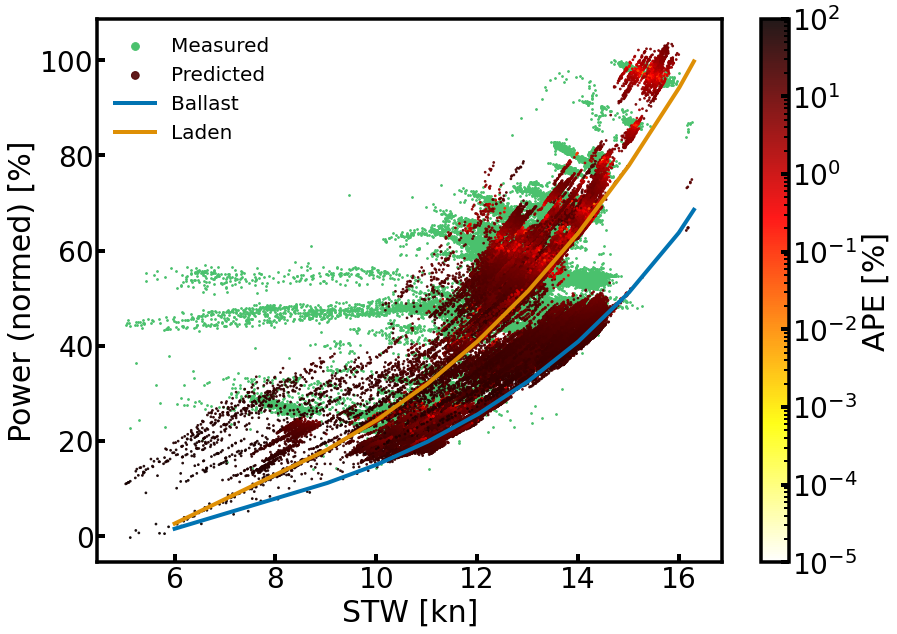}
\caption{Scatter plot of the measured and predicted main engine power (normalized w.r.t maximum) as a function of STW, together with the extrapolated laden and ballast sea trial curves. (Log-scaled color map.)}
\label{seatrial}
\end{figure}

\begin{figure}[!t]
\centering
\includegraphics[width=\linewidth]{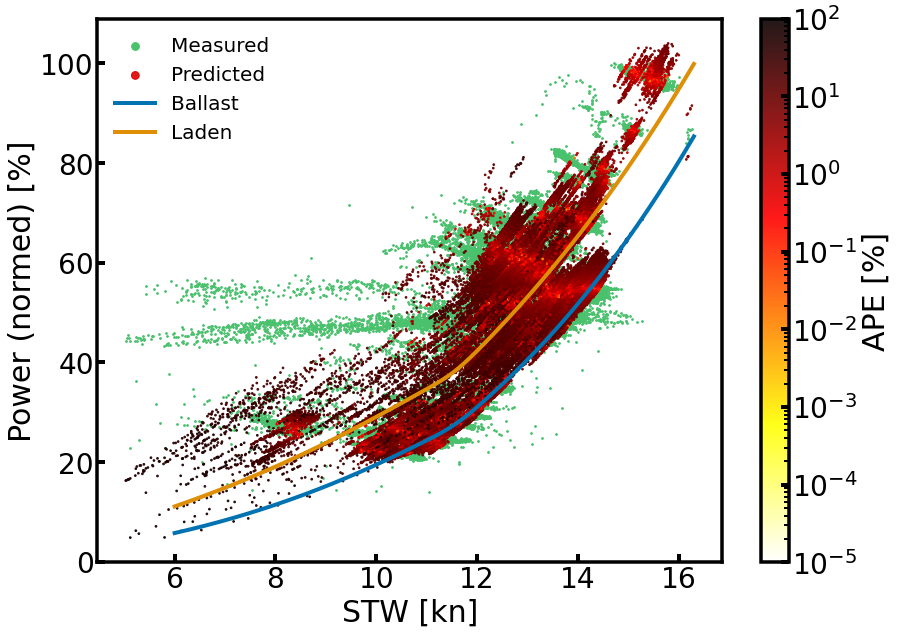}
\caption{Scatter plot of the measured and predicted main engine power (normalized w.r.t maximum) as a function of STW, together with the fitted calm-water laden and ballast curves. (Log-scaled color map.)}
\label{calmwater}
\end{figure}

Sea trial curves of a tanker were extrapolated to lower speeds, and weather corrections $R_{AA}$, $R_{AW}$ were calculated with Fujiwara's and Mittendorf's semi-empirical formulas, respectively. The main engine power is then obtained with Eq. \ref{eq:Pe}, ignoring $R_{AH}$ and $R_{others}$. Fig. \ref{seatrial} shows the predicted and measured power as a function of STW, along with the extrapolated laden and ballast sea trial curves. Predicted points $\hat{P}$ are colored based on their absolute percentage error (APE) = $|\hat{P} - P| / P$ with respect to the measured values $P$.
\\\\
The scatter plot in Fig. \ref{seatrial} supports our earlier hypothesis that extrapolating sea trial curves below the design speed leads to an underestimation of the power, indicating that the cubic law does not hold in this range. Additionally, time-dependent performance losses such as fouling can also cause an underestimation of the power, which increases over time.
\\\\
To address the limitations of the extrapolated sea trial curves, we perform a least-squares fit of Eq. \ref{eq:calm} on calm-water data according to the method by Berthelsen and Nielsen. The calm-water data is obtained by correcting the $H_S \leq 1$ m in-service power data for wind and waves with Fujiwara's and Mittendorf's formulas.\footnote{A histogram of $H_S$ can be found in Fig. \ref{app_Hs}.} First, the binary segmentation algorithm was used to find the breakpoint at $V_S = 11.53$ kn. Fig. \ref{calmwater} shows the results in the same way as before, but with the fitted calm-water curves. The speed-power exponent changes (smoothly ($\delta = 0.5$)) at the breakpoint from $\approx 1.8$ to $\approx 2.8$ for the laden curve.
\\\\
Table \ref{tab:calmbench} compares the accuracy metrics obtained with the sea trial curves and the fitted calm-water model. Our analysis shows that the data-driven calm-water method significantly improves the accuracy compared to extrapolated sea trial curves in terms of R2, MAPE, and MBE. This suggests that the data-driven calm-water model is a more reliable approach for predicting power requirements in realistic operating conditions. However, as shown in Fig. \ref{calmwater}, the fitted calm-water model still tends to underestimate the power (negative MBE) at low STW. This error is likely due to the presence of \emph{bands} in the speed-power data, where the vessel encounters heavy weather while operating in constant power or constant shaft rpm autopilot, resulting in large power values at low STW.
\begin{table}[!b]
\centering
\caption{Accuracy metrics evaluated on the predictions from different calm-water models in combination with Mittendorf's and Fujiwara's resistance corrections.}
\label{tab:calmbench}
\begin{tabular}{@{}lcccc@{}}
\toprule
Calm-Water Model & R2   & MAPE [\%] & MBE [kW] \\ \midrule
Extrapolated Sea Trial Curve       & 0.72    & 15     & -1522    \\
Fitted Calm-Water Curve      & 0.89  & 7.1     & -443     \\ \bottomrule
\end{tabular}
\end{table}

\mbox{}

\subsection{Added Wave Resistance Benchmark}
\begin{figure*}[!t]
\centering
\includegraphics[width=\linewidth]{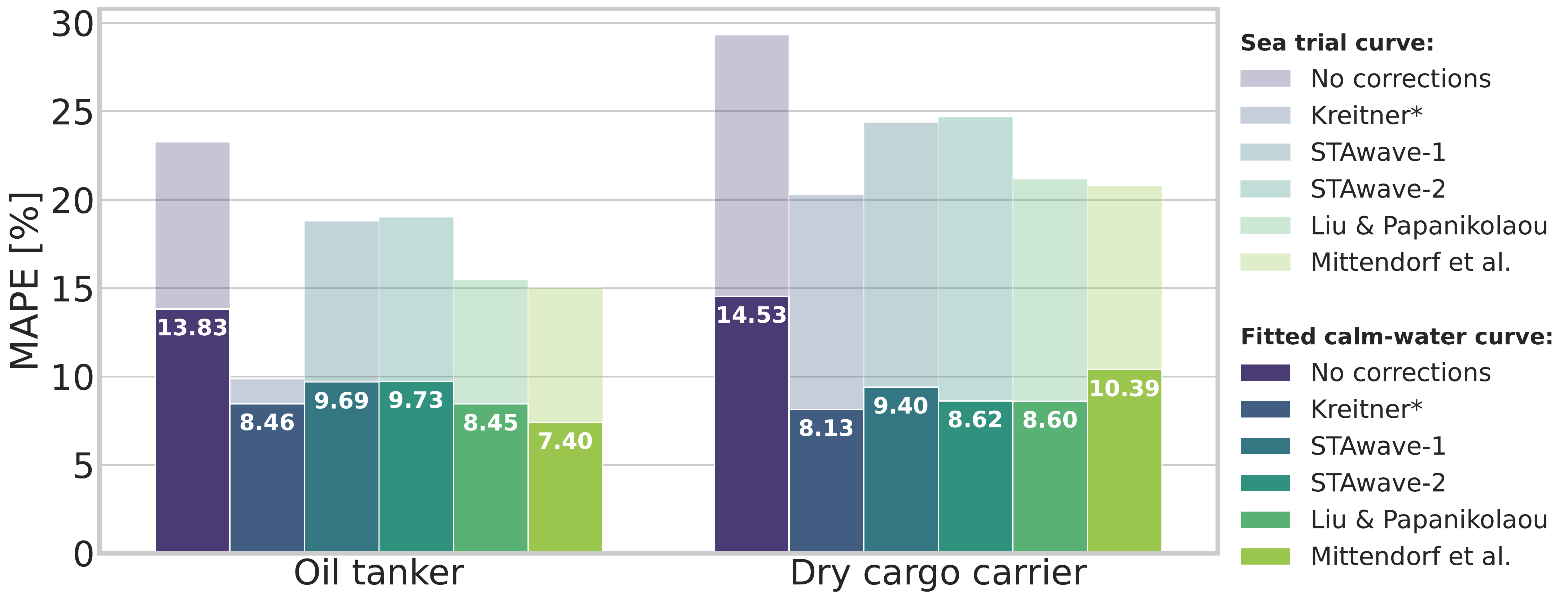}
\caption{MAPE obtained with the different combinations of wave resistance theories and calm-water models evaluated on multiple vessels.}
\label{benchmark}
\end{figure*}

Up to this point, we have only used Mittendorf's formula to calculate the power corrections due to waves. In the following analysis, we perform a benchmark of different added wave resistance theories to evaluate their performance, along with sea trials and the fitted calm-water regression model, using data from both an oil tanker and a dry cargo carrier. The method is similar to that used to produce Figs. \ref{seatrial} and \ref{calmwater}, except that different wave theories are used in combination with Fujiwara's $R_{AA}$. Fig. \ref{benchmark} shows the results, including the MAPE scores for each vessel and wave resistance theory, along with a comparison of the fitted calm-water model to sea trials. 
\\\\
Interestingly, we find that the accuracy of all the added resistance theories is very similar, despite significant differences in their complexity. One possible explanation for this finding is that the more advanced approaches are too complex for noisy data. When examining Fig. \ref{wavedrags}, we see that the Liu theory exhibits abrupt changes (non-differentiable points) in added wave resistance, which are not physically realistic. Based on these results, one might even consider using the wave angle-dependent Kreitner's formula, which has the advantage of requiring fewer ship-specific parameters and can be evaluated approximately ten times faster than Mittendorf's method ($\approx 0.1$ ms versus $\approx 1$ ms). 

\subsection{A Simple NN}
\label{sec:simple_nn}
\begin{figure}[!b]
\centering
\includegraphics[width=\linewidth]{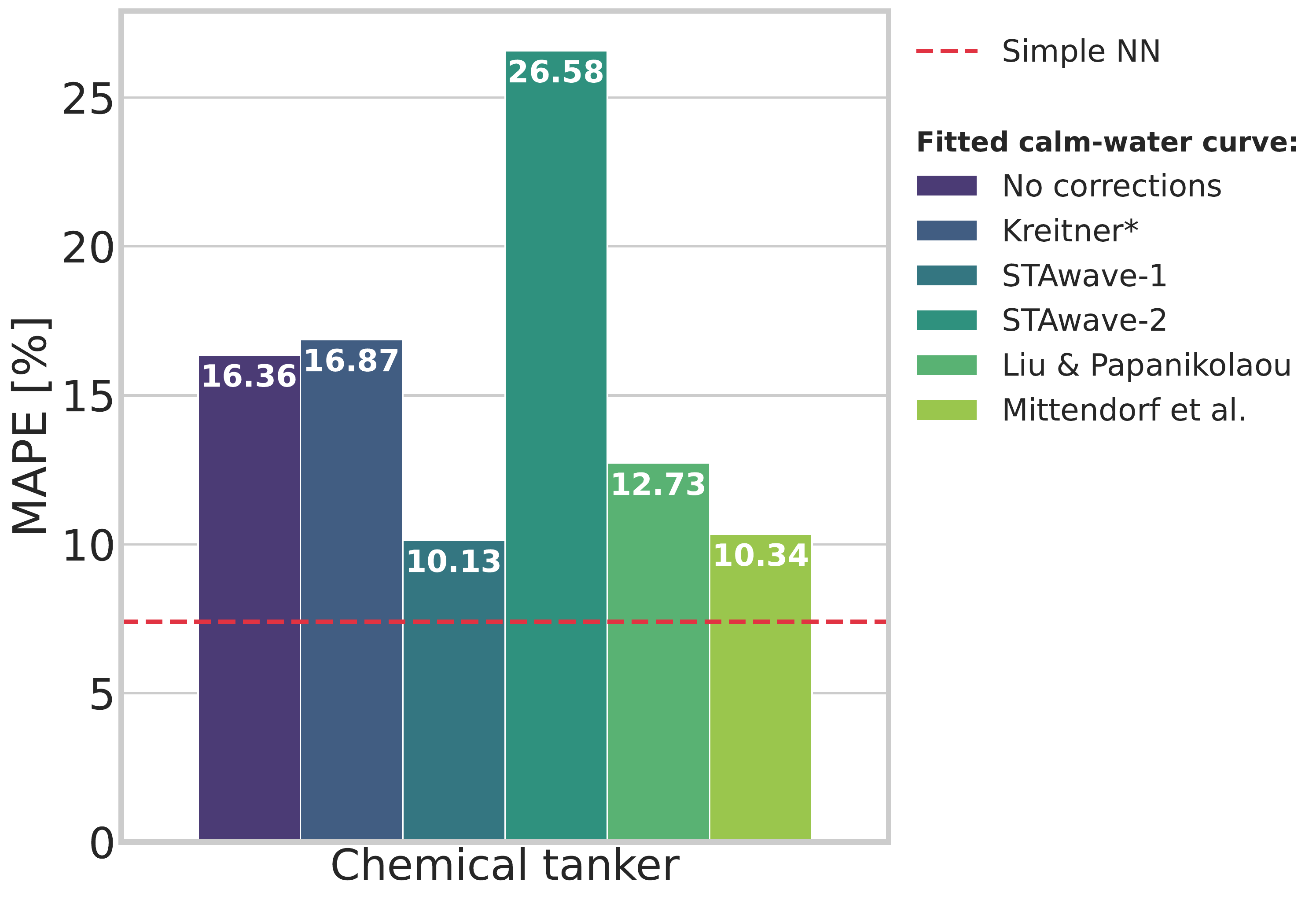}
\caption{MAPE obtained with the different combinations of wave resistance theories and the fitted calm-water model. A comparison is made with a simple NN.}
\label{MAPE_NN}
\end{figure}

\begin{figure}[!b]
\centering
\includegraphics[width=\linewidth]{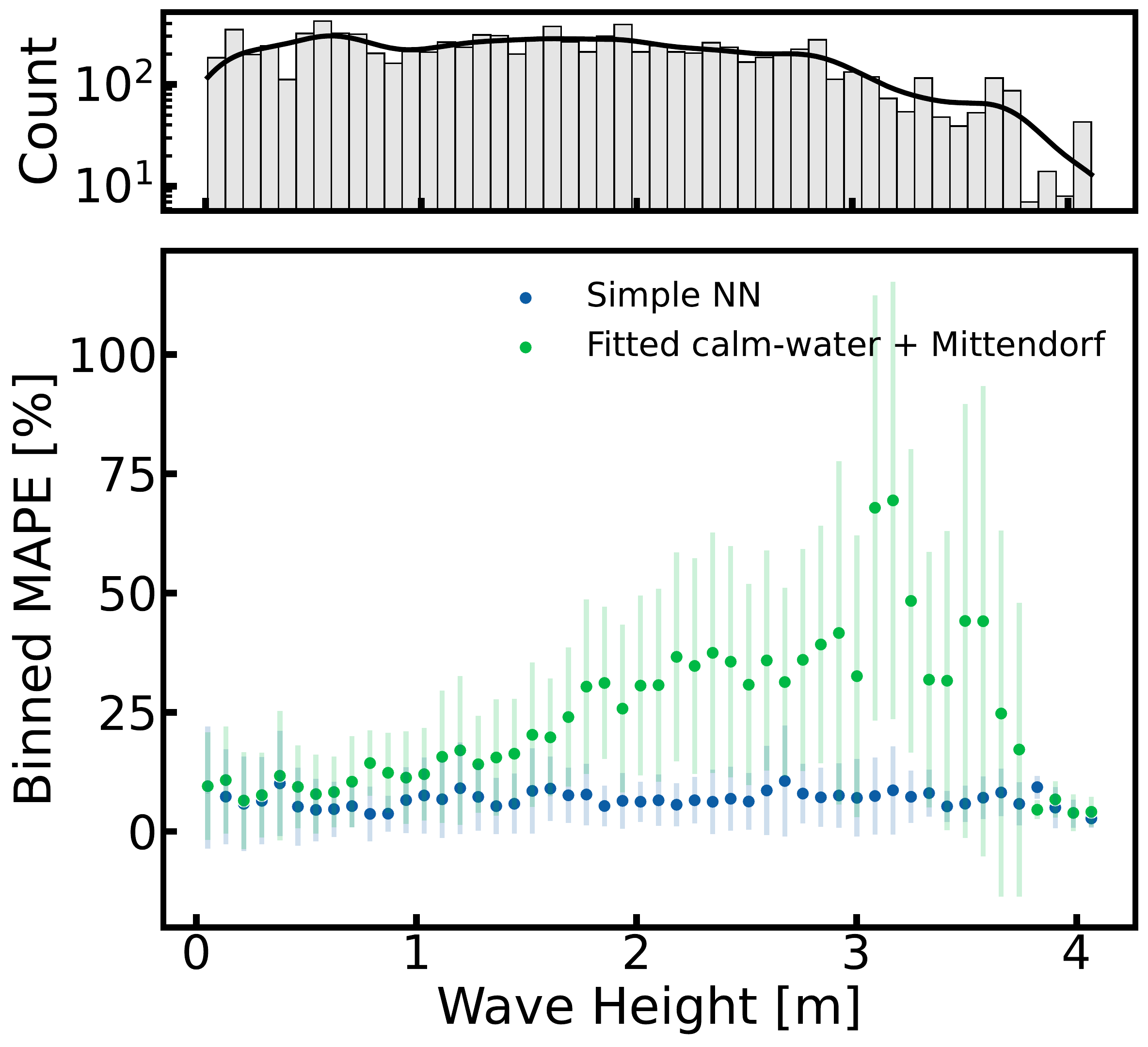}
\caption{MAPE binned w.r.t. the wave height, evaluated on a chemical tanker. A comparison is made between the fitted calm-water model in combination with Mittendorf's wave resistance and a simple NN. The histogram shows the data spread of the wave height.}
\label{binned_mape_wave}
\end{figure}

\begin{figure}[!b]
\centering
\includegraphics[width=\linewidth]{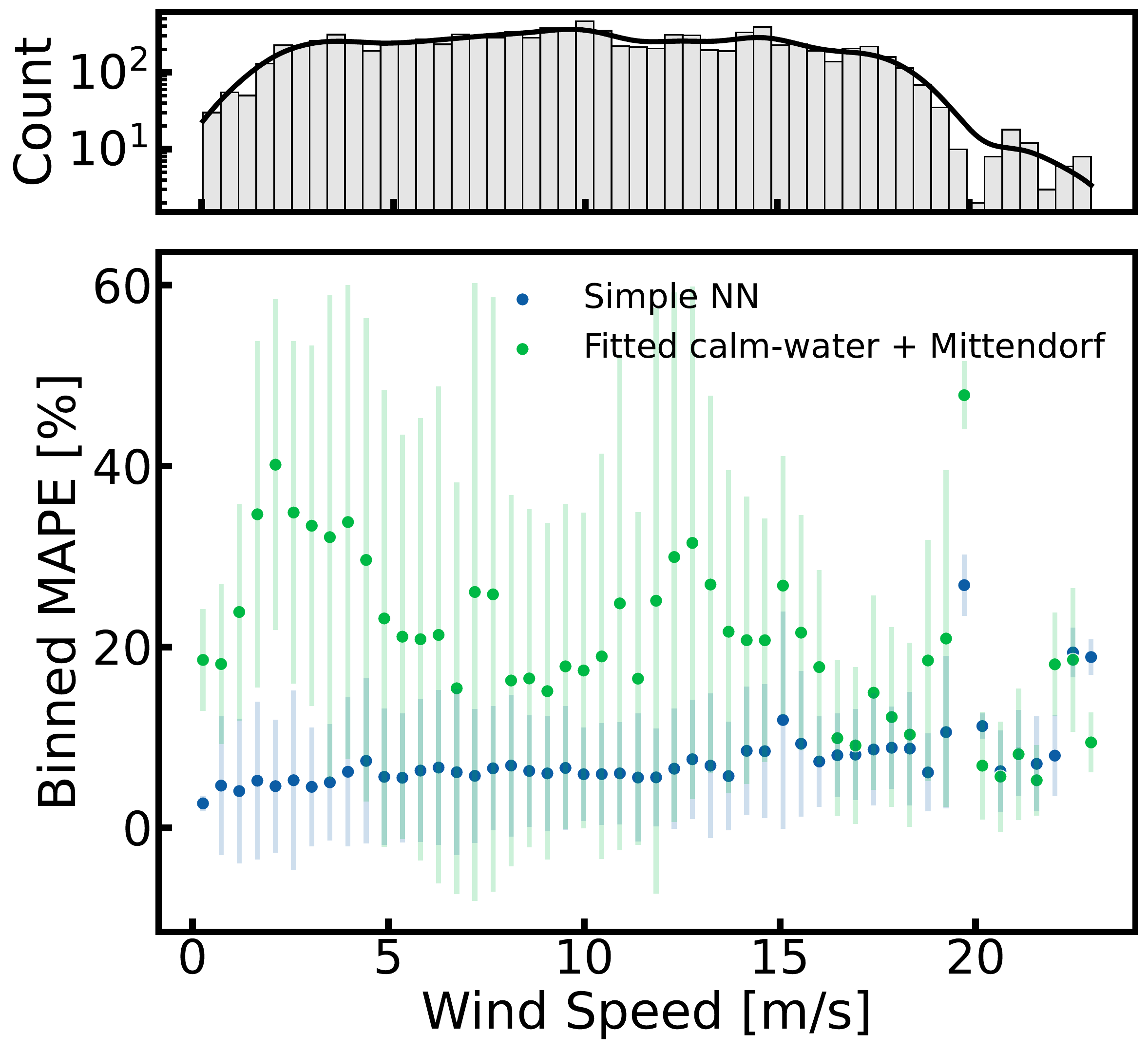}
\caption{MAPE binned w.r.t. the wind speed, evaluated on a chemical tanker. A comparison is made between the fitted calm-water model using Mittendorf's wave resistance and a simple NN. The histogram shows the data spread of the wind speed.}
\label{binned_mape_wind}
\end{figure}

In this section, we evaluate the performance of the simple NN by comparing it with the fitted calm-water model using different added wave resistance theories on the in-service data of a chemical tanker.\footnote{The FNN was trained using mini-batches of 32 samples with a learning rate of 0.015 for five epochs on each k-fold.} In contrast to the previous vessels, including sea trial curves in the analysis is not possible as they are unavailable. This further highlights the benefit of using the data-driven calm-water model by Berthelsen and Nielsen.
\\\\
Fig. \ref{MAPE_NN} compares the MAPE score achieved with the fitted calm-water curve using different wave resistance theories and the simple NN. The Kreitner and STAwave-2 theories perform worse than the model without weather corrections due to multiple data points falling outside the valid ranges for these empirical formulas. The simple NN achieves a MAPE of 7.4 \%, which represents a roughly 3 \% improvement over the best (semi-)empirical formula. This is a remarkable result considering the simplicity of our NN architecture. It is expected that further increasing the complexity and effort in developing the NN architecture would significantly improve the accuracy.
\mbox{}\\\\
To further analyze the performance of these models, we can perform an error analysis using binned scatter plots. In these plots, the data points are grouped into bins according to a variable of interest. Then, the MAPE is calculated for each bin, which allows us to visualize the relative accuracy of the model as a function of the variable of interest, providing insight into the factors that may influence the model's performance.
\\\\
Figs. \ref{binned_mape_wave} and \ref{binned_mape_wind} show a binned scatter plot with respect to the significant wave height and wind speed, respectively. Here, we compare the fitted calm-water model using Mittendorf's added wave resistance corrections with the simple NN from Fig. \ref{MAPE_NN}. The histograms show the data spread in the variable of interest, and the error bars denote the standard deviation of the binned MAPE. From both plots, we conclude that the NN achieves a better accuracy \emph{globally}, and thus it generalizes better than the theoretical model. 
\\\\
Taking the significant wave height as the variable of interest, the theoretical model performs particularly poorly in bad weather conditions, with a binned MAPE reaching over 70 \%. This effect is not as noticeable when evaluating the overall MAPE because the data spread gives increased weight to low wave heights. Similarly, using the relative wind speed as the variable of interest yields similar results for the NN, while the theoretical model performs worse for a wide range of wind speeds with a binned MAPE of up to 45 \%.
\mbox{}
\\\\
It is notable that, despite having less data available for high wave heights and wind speeds, the NN's accuracy remains approximately in a similar range. This finding is remarkable because NNs generally require considerable amounts of training data to learn and generalize effectively. The fact that the NN can maintain a similar level of accuracy with fewer data suggests that our choices of feature engineering were effective.

\subsection{Machine Learning as the Next Step}
\label{sec:next_step}
Our findings above demonstrate the potential of ML methods to improve the prediction of ship performance in various operating conditions compared to traditional methods. However, ML has previously been proposed as a valuable tool for shipping companies. Alexiou et al. \parencite*{alexiou2022towards} have compared a number of the latest data-driven models used in papers to model the speed-power relation and concluded that data-driven models were more suitable for shipping companies, provided sufficient representative historical data was available.
\\\\
The key aspect of this work lies in the direct comparison of the data-driven methods and multiple theoretical models on several vessels with different design types. However, before we propose ML as the natural next step for ship performance modeling, we must carefully weigh the benefits against the limitations.
\\\\
One of the main benefits of ML models is that they can be trained on existing data from a ship's performance, which means that they can be customized to a specific vessel and its operating conditions. This can provide more accurate and relevant results than using traditional methods.
\\
Another advantage of ML is that it can automate some of the more tedious and time-consuming aspects of ship performance modeling, freeing engineers to focus on other tasks. Machine learning algorithms can also learn from data in real-time, which means that they can adapt to changes in a ship's operating conditions and provide updated predictions on an ongoing basis.
\\\\
In contrast, one of the main constraints is the need for substantial amounts of high-quality data to train the models. This may be challenging for some ships or operating conditions where data may be limited or difficult to collect (e.g., only noon report data available or sensor drift).
\\\\
Additionally, ML models can be difficult to interpret and explain, making it challenging for engineers to apply the results to their work. This issue can be addressed through the use of explainable artificial intelligence (XAI) techniques \cite{gunning2019xai} and physics-informed ML (PIML) \cite{karniadakis2021physics}. XAI helps to make ML algorithms more transparent and interpretable, while PIML allows ML methods to follow physical relations.
\\\\
Furthermore, ML algorithms can be subject to bias if the training data is not representative of the full range of conditions that a ship may encounter. This can lead to inaccurate or misleading results, which can be challenging to detect and correct. However, careful feature engineering and the use of PIML can help to ensure that the ML algorithm follows physical relationships beyond the range of the training data and is not biased by the data used to train it. This can help to improve the robustness and generalizability of the ML model \cite{julie}.
\\\\
Overall, while the use of ML in ship performance modeling offers many potential benefits, there are also significant limitations that one must consider. However, as the availability of data increases and the expertise in ML continues to grow, these restrictions are becoming easier to overcome.

\section{Conclusion}
\label{sec:conclusion}
We have outlined several (semi-)empirical approaches developed in the last decades of ship hydrodynamics. Most of these methods use theoretical approximations, after which they are fitted to experimental data. Researchers have been using increasingly advanced regression algorithms to enhance the experimental fits of their semi-empirical formulas. 
\\
Although their findings are crucial to obtaining a better understanding of ship performance in a research-based way, the developed methods might not be the best choice in practical applications as they require extensive ship particulars that are often unavailable.
\\\\
Our benchmark of different added wave resistance theories, performed on in-service data from multiple vessels, supports this idea. Results showed that simple theories had similar accuracy to more complex methods, and that a data-driven calm-water resistance regression method proposed by Berthelsen and Nielsen \parencite*{berthelsen2021prediction} was more useful than extrapolated sea trial curves for performance monitoring.
\\\\
A natural next step to further improve accuracy and reduce the need for empirical methodologies and ship particulars, may be to leverage the data scalability of ML techniques. By applying ML techniques to ship-specific sensor data, the whole ship and its environment can be \emph{learned} directly from real-world data. 
\\
Our results showed that a simple NN outperforms all of the (semi-)empirical added wave resistance methods, which were used in combination with the data-driven calm-water model. This suggests that ML techniques may be a valuable tool for improving ship performance modeling in practical applications. Even a small increase in accuracy could have significant economic and ecological benefits for the shipping industry. It requires a tiny change of perspective -- transitioning from theoretical regression analysis to a more advanced approach -- potentially yielding significant benefits. 
\printbibliography 

\newpage
\appendix
\begin{figure}[!h]
\centering
\includegraphics[width=\linewidth]{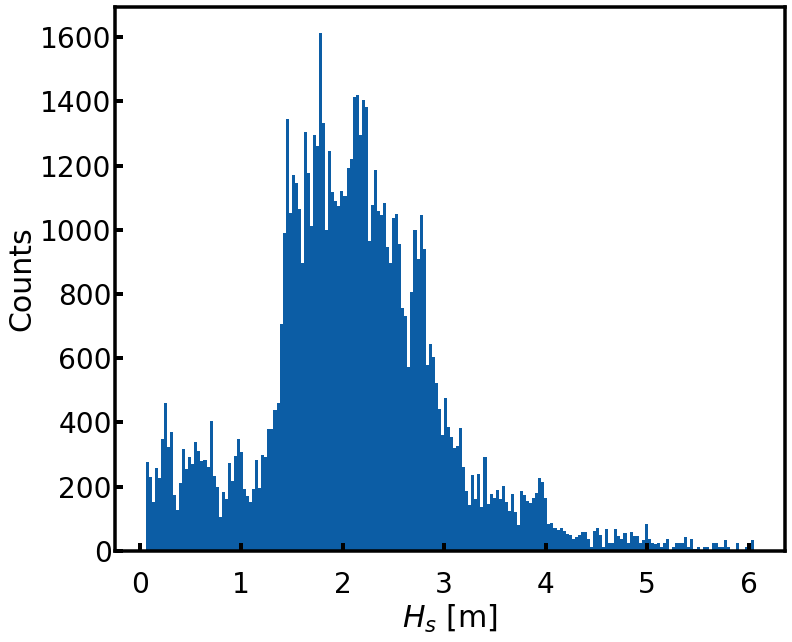}
\caption{Histogram of wave height.}
\label{app_Hs}
\end{figure}

\end{document}